\documentclass[times, twoside]{henriques-style}
\usepackage{blindtext}
\usepackage[nolist]{acronym}
\usepackage{amsmath}
\usepackage{amsfonts} 
\setcitestyle{square}
\usepackage{tikz}
\usetikzlibrary{positioning, calc, matrix, decorations.pathreplacing, shapes.geometric}
\usepackage{adjustbox}
\usepackage{xspace}
\makeatletter
\DeclareRobustCommand\onedot{\futurelet\@let@token\@onedot}
\def\@onedot{\ifx\@let@token.\else.\null\fi\xspace}

\leadauthor{Chung} 

\begin{document}

\title{Domain-Robust Mitotic Figure Detection with Style Transfer}
\shorttitle{MIDOG CGV Algorithm}
% Use letters for affiliations, numbers to show equal authorship (if applicable) and to indicate the corresponding author
\author[1]{Youjin Chung\thanks{Y. Chung and J. Cho contributed equally to this work as first authors}}
\author[1]{Jihoon Cho}
\author[1]{Jinah Park}

\affil[1]{Computer Graphics and Visualizations lab., School of Computing, Korea Advanced Institute of Science and Technology, Daejeon, South Korea}

\maketitle
\begin{figure}[h]
    \centering
    \includegraphics[width=\linewidth]{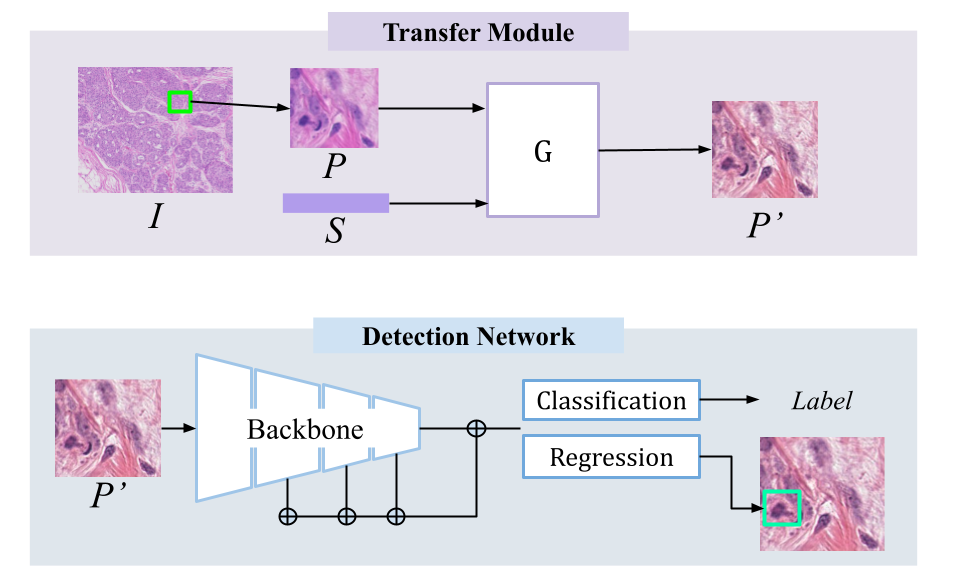}
    \caption{Overall architecture of our model. I: \acl{wsi}, P: patch, \\ S: 4-component style code, G: style transfer module, P': style transferred patch}
    \label{fig:arch}
\end{figure}
%TC:break Abstract
%the command above serves to have a word count for the abstract
\begin{abstract}
We propose a new training scheme for domain generalization in mitotic figure detection. Mitotic figures show different characteristics for each scanner. We consider each scanner as a 'domain' and the image distribution specified for each domain as 'style'. The goal is to train our network to be robust on scanner types by using various 'style' images. To expand the style variance, we transfer a style of the training image into arbitrary styles, by defining a module based on StarGAN. Our model with the proposed training scheme shows positive performance on MIDOG Preliminary Test-Set containing scanners never seen before.

\end {abstract}
%TC:break main
%the command above serves to have a word count for the abstract

%\begin{keywords}
%Domain Generalization | Mitotic Count | Histopathology | Style Transfer
%\end{keywords}

\begin{corrauthor}
jinahpark@kaist.ac.kr
\end{corrauthor}

\section*{Introduction}
\acl{mc} of histological tumor images is important in digital pathology. Due to the  high cost of detecting miotic figures, auto-detection idea has gained more attention these days. The challenge of the \acl{midog} (\acs{midog}) \cite{ref_midog_desc} is to detect mitotic figures in \acl{wsi} (\acs{wsi}) from different scanners. \acs{midog} challenge data consists of 200 images from four different scanners (50 images for each) with 150 labels. We consider the characteristics of the image like hue and contrast as a ‘style’ in \acs{wsi}, without morphological features. Then we see that each scanner differs in its style, which also makes the difference in data distribution. As a reference algorithm \cite{ref_midog}, \acs{midog} used the RetinaNet~\cite{ref_retina} a widely-used supervision-based detection network. However, the nature of the supervision leads to weak performance in unseen data distributions. To overcome this domain dependency problem in scanners, we transfer the style in variety for the generation of differently styled training images using StarGAN \cite{ref_style} which can shift the input between multiple domains. In this way, the supervised detection model learns from different styles, eventually detecting the mitotic figures in the images from an unseen scanner.  

\section*{Method}
\begin{figure}[h]
    \centering
    \includegraphics[width=\linewidth]{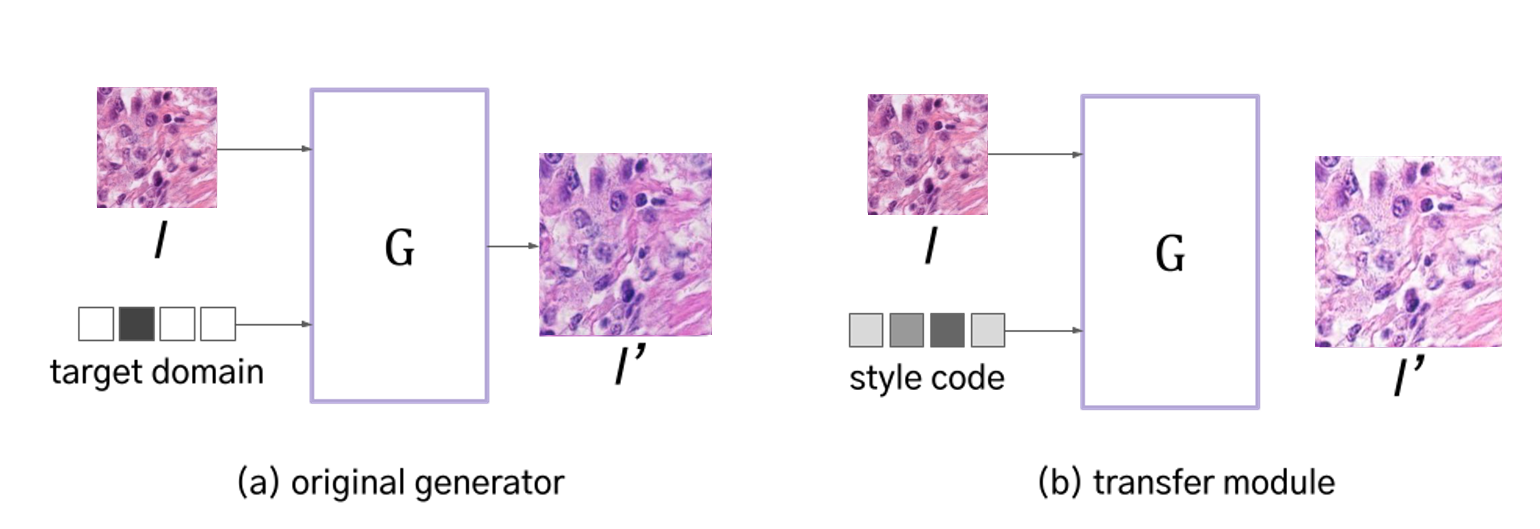}
    \caption{Using StarGAN's generator as transfer module: (a) original generator operation, (b) transfer module operation with style code}
    \label{fig:transfer}
\end{figure}

Our goal is to train the model with various domains because supervision from a set of data from the limited scanners hinders the model from detecting a target in new scanner images. To be robust with scanner vendors, the generator is trained prior to the detection network to translate the image into a specific scanner style as Fig.\ref{fig:transfer} (a). We use the generator as a style transfer module which generates a variously styled image based on the mixing of the scanner characteristics as Fig.\ref{fig:transfer} (b). The overview of our approach is described in Fig.\ref{fig:arch}. The transfer module generates a new styled patch \textit{P'} from the original patch \textit{P} with style code \textit{S}. Then the detection network is trained using \textit{P'} for the mitotic figure detection task.
\subsection*{Networks}
As to decide our base-network for detection of mitotic figures, we conducted a pre-test on many well-known detection networks such as RetinaNet~\cite{ref_retina}, CenterNet~\cite{ref_center}, CenterNet2~\cite{ref_center2}, etc. From the pre-test, we found out the RetinaNet fits the best for our detection network. This is because large and complex models are presumed to be overqualified for our task. We also tested on the size of the backbone of our model, which is ResNet-50 and ResNet-101 each, ResNet-101 showed a little bit better result. \\

To create images of different styles, we used the generator of StarGAN~\cite{ref_style}. We refer it as the transfer module. This module guarantees transferring only the style without losing morphological information such as structures of a cell and nucleus upon training, explained in the next section. In training, we generate different-styled images using a random style code for the diversity of styles. This style code is a 4-component normalized vector representing the contribution of each original patch style to be transferred to the module. 
\subsection*{Training Scheme}
Training procedure of our detection network consists of two stages: first, training the transfer module, and then training the detection network using the trained transfer module.
\begin{figure}[h]
    \centering
    \includegraphics[width=0.95\linewidth]{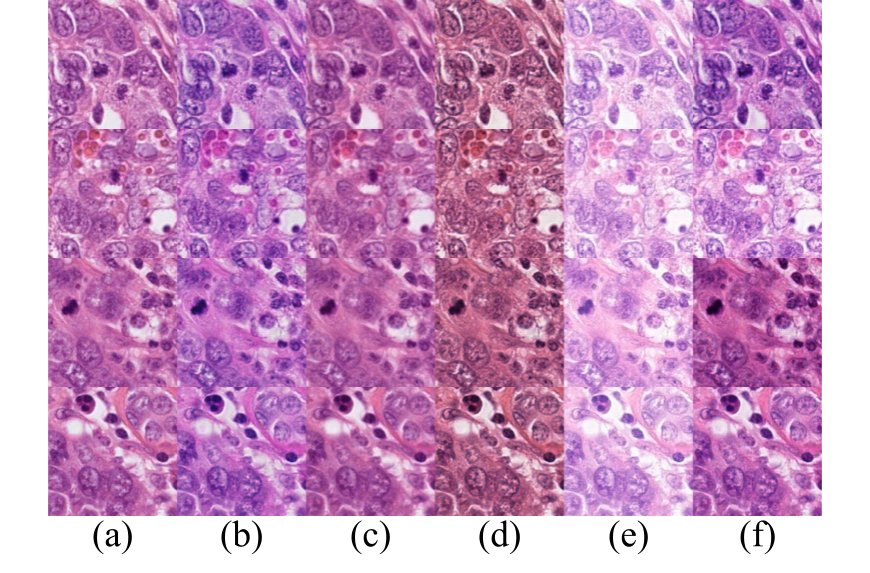}
    \caption{Transfer module results. (a): original patch (b): scanner-1 style \\ (c): scanner-2 style (d): scanner-3 style (e) scanner-4 style (f): random style}
    \label{fig:generator}
\end{figure}
\subsubsection*{Transfer Module Training} \label{sec:train_transfer}
We followed the adversarial training scheme and loss function of Choi et al~\cite{ref_style} with four scanner images provided. Each scanner type represents a one-hot label of four-component style code vector \textit{S}. In the training step the generator \textit{G} is trained to transfer one scanner image into four scanner's styled images, while the discriminator \textit{D} is trained to discriminate whether the image is synthesized or not. For domain-wise classification loss (equation \ref{eq:cls}), $c \in [0, 1, 2, 3]$ is used to transfer between four scanners. Also, for preventing the losing morphological features a reconstruction loss (equation \ref{eq:rec}) is applied, which restores the original image from the style transferred image. By this approach, \textit{G} will learn to transfer between scanner styles, without losing the morphology. The result of the transferred patch generated by \textit{G} is shown in Fig.\ref{fig:generator}. The final loss of both generator and discriminator is shown in equation (\ref{eq:3}). Finally, after training, \textit{G} of GAN network is used as the transfer module in our model, using random style code as (f) in Fig.\ref{fig:generator}.

\begin{align}
    \begin{split}\label{eq:cls}
        L_{cls}^r &= \mathbb{E}_{x, c'}[-\log D_{cls}(c'|x)] \\
        L_{cls}^f &= \mathbb{E}_{x, c}[-\log D_{cls}(c|G(x, c))]        
        \end{split} \\[1em]
        \label{eq:rec}
        L_{Rec} &= \mathbb{E}_{x, c, c'}[||x - G(G(x, c), c')||_1] \\[1em]
        \begin{split}\label{eq:3}
        L_D &= -L_{adv} + \lambda_{cls}L_{cls}^r, \\
        L_G &= L_{adv} + \lambda_{cls}L_{cls}^f + \lambda_{rec}L_{rec}        
    \end{split}
\end{align}    
\vspace{0.5em}
\subsubsection*{Detection Network Training} The resolution of WSI is too high to detect the mitotic figure at once due to computational overload, so we chose the patch-based approach for both training and inference. However, training with whole patches might cause a high imbalance problem between foreground patches (including ground truths) and background patches (not including ground truths), we have adjusted the ratio of the foreground and background patches experimentally. We adjusted anchors $A_{b}$ of equation \ref{eq:anchor}, from $[32, 64, 128, 256, 512]$ to $[50, 50, 50, 50, 50]$ because the size of the bounding box is always 50. To resolve the imbalance problem in the scanner's foreground patch counts, we used an uniform number of training patches from each scanner. Then for each patch, we used the pre-trained transfer module to change the scanner style into an arbitrary scanner style using a random 4-component style code as shown in Figure \ref{fig:transfer}. We used arbitrary styled image with a certain probability $p$, and original image with probability $p-1$. This is because the transfer module cannot generate the pathological feature-preserved image perfectly and over-generalization can degrade the performance. The training loss functions are same of Lin at al~\cite{ref_retina} using focal loss (\ref{eq:focal}). On the inference step, all patches from \acs{wsi} are used without using style transfer, and the detection result of a mitotic figure is aggregated with non-maximum suppression (NMS).
\begin{equation}\label{eq:anchor}
    A = [x \times 2^0, x \times 2^{\frac{1}{3}}, x \times 2^{\frac{2}{3}}],  \forall x \in A_{base}
\end{equation}
\begin{equation}\label{eq:focal}
    FL(p_t) = -\alpha_t(1 - p_t)^\gamma log(p_t)
\end{equation}

\subsection*{Implementation and Preliminary Test-set Results}
Our detection network was built on detectron2 \cite{wu2019detectron2}, used RetinaNet implementation. Transfer module used the official code in \cite{ref_style}. The training was held on NVIDIA TITAN RTX for 1.6M, 500K iterations for transfer module and detection network. During implementation, we used bg\_fg ratio $\alpha=6$ with random style patch portion $p=0.2$, set the detection score confidence as 0.7 and random rotation augmentation $R \in [0, 90, 180, 270]$. Learning rate starts from 0.2 and drops $\times0.1$ for steps (320K, 450K) F1 score is used as our evaluation metric same as used in \acs{midog}, which uses both precision and recall of Mitotic Figures' detection results. Finally, Our model achieved F1 score of 0.7548, recall of 0.7048 and precision of 0.8125 in the preliminary test phase. 

\begin{acronym}
\acro{mc}[MC] {Mitotic Count}
\acro{midog}[MIDOG]{MItosis DOmain Generalization}
\acro{miccai}[MICCAI]{Medical Image Computing and Computer Assisted Intervention}
\acro{wsi}[WSI]{Whole Slide Image}
\acro{he}[H\&E]{Hematoxylin \& Eosin}
\acro{grl}[GRL]{Gradient Reverse Layer}
\acro{aucpr}[AUCPR]{area under the precision-recall curve}
\end{acronym}

\end{document}